\documentclass[conference]{IEEEtran}
\IEEEoverridecommandlockouts
\usepackage{cite}
\usepackage{amsmath,amssymb,amsfonts}
\usepackage{algorithm}
\usepackage{algpseudocode}
\usepackage{graphicx}
\usepackage{textcomp}
\usepackage{xcolor}
\usepackage{hyperref}

\def\BibTeX{{\rm B\kern-.05em{\sc i\kern-.025em b}\kern-.08em
    T\kern-.1667em\lower.7ex\hbox{E}\kern-.125emX}}
\begin{document}

\title{Multi-Scale Diffusion: Enhancing Spatial Layout in High-Resolution Panoramic Image Generation}


\author{
     Xiaoyu Zhang$^{1}$, 
     Teng Zhou$^{1}$, 
     Xinlong Zhang$^{1}$, 
     Jia Wei$^{1}$, 
     Yongchuan Tang$^{1}$\textsuperscript{*}\thanks{This work was supported by the Lingyan Plan of Zhejiang Province under Grant No. 2025C02211.} \thanks{*Yongchuan Tang is the corresponding author.} \\
     \emph{$^{1}$ Zhejiang University, Hangzhou, China} \\ 
     {\small \tt \{xiaoyzhang,tengzhou,xinlzhang,weijia\_77,yctang\}@zju.edu.cn}
}

\maketitle

\begin{abstract}
    Diffusion models have recently gained recognition for generating diverse and high-quality content, especially in image synthesis. These models excel not only in creating fixed-size images but also in producing panoramic images. However, existing methods often struggle with spatial layout consistency when producing high-resolution panoramas due to the lack of guidance on the global image layout. This paper introduces the Multi-Scale Diffusion (MSD), an optimized framework that extends the panoramic image generation framework to multiple resolution levels. Our method leverages gradient descent techniques to incorporate structural information from low-resolution images into high-resolution outputs. Through comprehensive qualitative and quantitative evaluations against prior work, we demonstrate that our approach significantly improves the coherence of high-resolution panorama generation.
\end{abstract}

\begin{IEEEkeywords}
    Diffusion Model, Panoramic Image Generation, High-Resolution, Gradient Descent
\end{IEEEkeywords}

\section{Introduction}
\label{sec:intro}

    Recent advances in diffusion models have shown remarkable capabilities in image synthesis \cite{ho2020denoising, dhariwal2021diffusion, song2020denoising}. These models employ dual Markov chains to learn data distributions through simulated diffusion and denoising processes. The generated images demonstrate superior quality compared to those synthesized by other generative approaches \cite{kobyzev2020normalizing, karras2019style, van2017neural}. Models like Stable Diffusion \cite{rombach2022high, podell2023sdxl}, trained on large and diverse datasets, have significantly advanced the field. These models demonstrate exceptional capability in generating detailed, contextually appropriate images, emerging as fundamental components of generative AI with broad applications.

    Panoramic image generation \cite{BarTal2023MultiDiffusionFD, feng2023diffusion360, zhou2024holodreamer} produces images with variable aspect ratios. This technology provides an expansive field of view, enhancing visual completeness and immersion. Despite significant research attention, this field faces several challenges, particularly the limited availability of training data. The scarcity of data hinders diffusion models from directly generating panoramic images.

\begin{figure*}[t]
    \centering
    \includegraphics[width=1\textwidth]{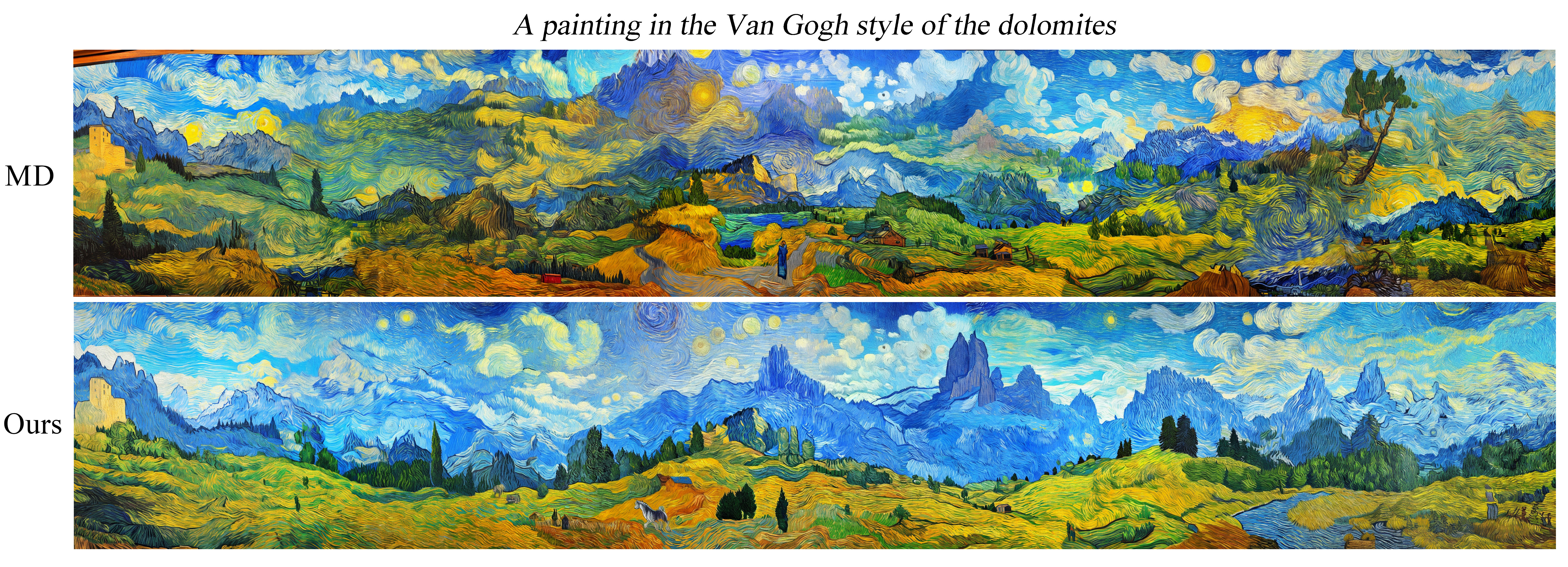} 
    \caption{Comparison of high-resolution panoramic images generated by MultiDiffusion and our Multi-Scale Diffusion. MultiDiffusion excels in seamless image stitching but struggles with spatial coherence, leading to structural inconsistencies. In contrast, our model improves on this by integrating coarse structures and fine details from different resolution levels, producing panoramas that are both structurally coherent and visually detailed. }
    \label{fig:1}
    \end{figure*}

    Existing methods stitch together images generated by multiple diffusion models to tackle these challenges. These methods fall into two categories: image extrapolation \cite{avrahami2022blended, avrahami2023blended} and joint diffusion \cite{zhang2023diffcollage, tang2023emergent, jimenez2023mixture}. The first approach involves generating the final output by extrapolating the edges of an initial image. However, this method frequently produces repetitive patterns, leading to unrealistic panoramas. Joint diffusion has become the leading method for creating seamless panoramic images. This approach integrates models with shared parameters or constraints, blending noisy images across overlapping regions by averaging the intermediate outputs at each denoising step. MultiDiffusion (MD) \cite{BarTal2023MultiDiffusionFD} exemplifies this approach and significantly advances panoramic image generation. However, the MD framework has limitations in generating high-resolution panoramas. The absence of global layout guidance leads to a disorganized spatial arrangement, compromising the final image quality (Fig. \ref{fig:1}). 

    We introduce Multi-Scale Diffusion (MSD) as a solution to the challenges of generating high-resolution panoramic images. The method combines structure and composition from lower-resolution images with details from higher resolutions. The single-step denoising process is divided into multiple stages, using a phased approach that gradually improves the panoramic image's quality. The framework extends the MultiDiffusion technique for joint denoising at each resolution level. MSD employs low-resolution images as structural guides through gradient descent optimization, effectively minimizing inconsistencies across resolution layers. As shown in Fig.\ref{fig:1}, our method successfully generates panoramic images with large-scale coherence and fine-grained detail.

    The main contributions of our work are as follows:
    \begin{itemize}
        \item We identify integrating global layout information as essential for producing high-quality high-resolution panoramic images.
        
        \item Building on our theoretical findings, we propose the Multi-Scale Diffusion (MSD) framework. Our method incorporates spatial guidance via gradient descent, producing panoramas that are both structurally coherent and rich in detail.
        
        \item Comprehensive experiments validate the effectiveness of our approach, which surpasses baselines in both quantitative metrics and qualitative evaluations, solidifying it as a superior solution for panoramic image generation.
    \end{itemize}

\section{Related Work}
    \subsection{Diffusion Models}
        Diffusion models \cite{sohl2015deep, dhariwal2021diffusion, song2020score, nichol2021improved} represent a class of generative models that create data through a two-step process: forward diffusion and reverse denoising. This process uses Markov chains to gradually perturb and reconstruct data distributions, enabling data creation from noise step-by-step. 
        
        The field has advanced considerably since the emergence of DDPM \cite{ho2020denoising}. These models have demonstrated remarkable success in image generation tasks \cite{saharia2022photorealistic, dhariwal2021diffusion}, outperforming previous approaches like GANs \cite{karras2020analyzing, karras2019style}, VAEs \cite{kingma2013auto, van2017neural} and Flows \cite{kobyzev2020normalizing}. DDIM \cite{song2020denoising} further improved the process by introducing non-Markovian transitions that predict denoised data, significantly accelerating the denoising process. LDMs \cite{rombach2022high} marked another breakthrough by implementing the diffusion process in latent space using a pre-trained autoencoder, leading to high-performance systems like Stable Diffusion \cite{rombach2022high, podell2023sdxl} and DALLE2 \cite{ramesh2022hierarchical}. 
        
    \subsection{Panoramic Image Generation}

        The use of diffusion models for panoramic image generation has garnered significant research attention. Due to challenges in data acquisition and computational efficiency, directly generating panoramas presents numerous difficulties. Existing methods can be mainly divided into two categories: image extrapolation methods \cite{avrahami2022blended, avrahami2023blended}, which extrapolate image edges, and approaches that integrate multiple diffusion paths to fuse overlapped denoising paths without additional training or fine-tuning \cite{jimenez2023mixture, zhang2023diffcollage, tang2023emergent}. 
        
        The MultiDiffusion \cite{BarTal2023MultiDiffusionFD} exemplifies the second category of diffusion models, proving feasible and practical. It has laid a foundation for numerous subsequent studies. SCALECRAFTER \cite{he2023scalecrafter} adapts diffusion models for higher resolutions by dilating convolution kernels, while Demofusion uses an "upsample-diffuse-denoise" loop to generate high-resolution ones progressively. However, these methods focus on generating high-resolution images rather than panoramic ones. SyncDiffusion \cite{lee2023syncdiffusion} ensures consistency across panoramic images by calculating perceptual loss across windows. TwinDiffusion \cite{zhou2024twindiffusion} achieves smoother transitions by aligning adjacent image parts. Nevertheless, these methods are limited to generating scene images repetitively in either the length or width direction. When extending in both directions simultaneously, conflicting scene layouts in different windows result in a chaotic overall image layout.

        To address this issue, we build upon existing methods by incorporating guidance from low-resolution images to capture structural details. This approach enables the generation of panoramas that are both rich in detail and structurally sound.

\begin{figure*}[t]
    \centering
    \includegraphics[width=1\textwidth]{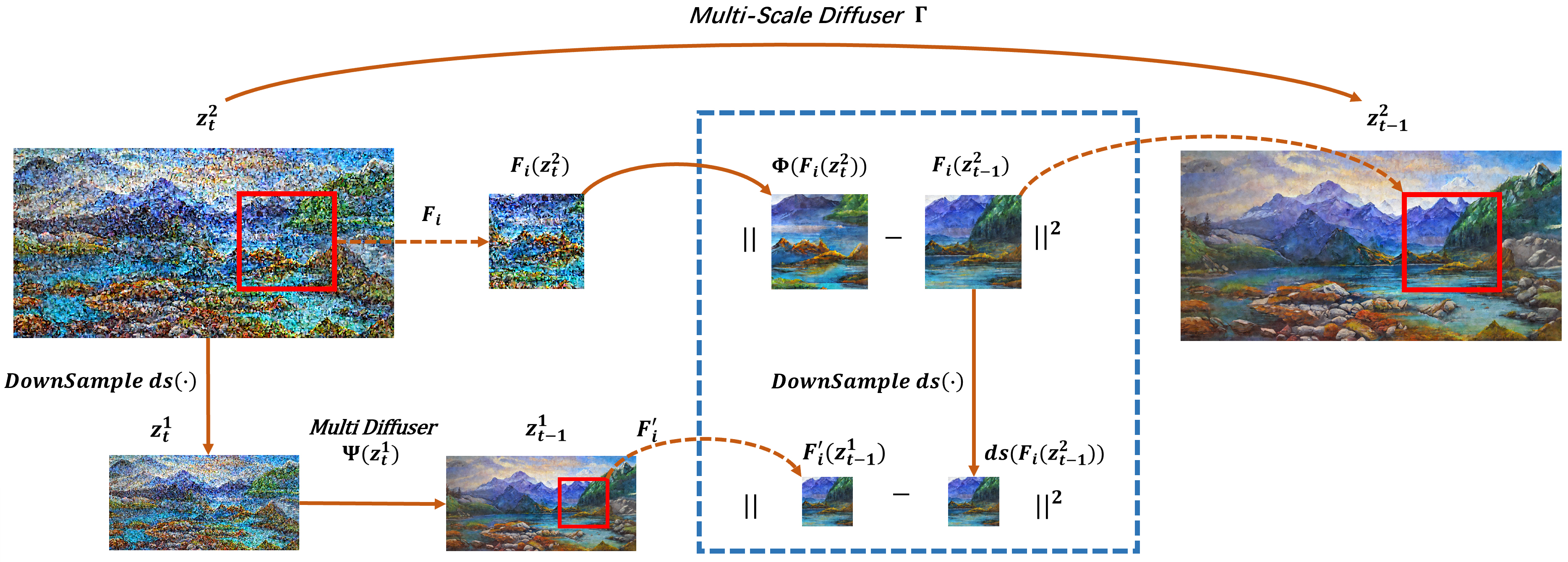} 
        \caption{Our MSD framework. The single-step denoising process is straightforward in multiple stages, progressively denoising the panoramic image from low to high resolution. Building on the pre-trained model, denoted as $\Phi$, a new generation method, called $\Gamma$, is introduced. This method uses the results from the low-resolution denoising as constraints to guide the high-resolution denoising. The optimization objectives are twofold: (i) ensuring the final denoised image $F_i\left(\Gamma\left(z_t^2\right)\right)$ closely matches the denoised results of each cropped window $\Phi({F_i(z}_t^2))$; (ii) maintaining consistency between denoised images across different resolution layers ($z_{t-1}^2$, $z_{t-1}^1$). For the lowest-resolution image, our method simplifies to the standard MultiDiffusion approach, denoted by $\Psi$.}
    \label{fig:2}
\end{figure*}

\section{Method}
    \subsection{Preliminary}
        \subsubsection{Latent Diffusion Model}
            We introduce a pre-trained diffusion model operating in a latent space $\mathbb{R}^{c\times h\times w}$. The model generates image $x_0$ through iterative denoising, starting with initial Gaussian noise $x_T$. This process follows a predefined noise schedule, updating the current image $x_t$ at each timestep $t$ with the following formula:
    
            \begin{equation} 
                \begin{aligned} 
                x_{t-1} &= \sqrt{\alpha_{t-1}}\left(\frac{x_t-\sqrt{1-\alpha_t}\epsilon_\theta\left(x_t,t\right)}{\sqrt{\alpha_t}}\right)+ \\
                        &\quad \sqrt{1-\alpha_{t-1}}\epsilon_\theta\left(x_t,t\right),  
                \end{aligned}
            \end{equation}
            
            \noindent 
            where $\alpha_t$ is parameterized by the noise schedule, $\epsilon_\theta\left(x_t,\ t\right)$ is the noise predicted by the denoising model at timestep $t$, parameterized by $\theta$. For brevity, we denote the denoising steps as $\Phi$ in the rest of the paper:
            
            \begin{equation} 
            \label{eq:sd}
                x_{t-1}=\ \Phi\left(x_t\right),
            \end{equation}
        
        \subsubsection{MultiDiffusion}
            The MultiDiffusion \cite{BarTal2023MultiDiffusionFD} framework extends LDMs \cite{rombach2022high} by employing a multi-window joint diffusion technique. In this approach, the denoising process of the model $\Psi$ is conducted a latent space $\mathbb{R}^{c\times H\times W}$ with $H>h$ and $W>w$. Initially, the panoramic image $z_t\in\mathbb{R}^{c\times H\times W}$ is cropped into a series of window images:
            
            \begin{equation} 
            \label{eq:crop}
                x_t^i=F_i\left(z_t\right),
            \end{equation}

            \noindent 
            where $F_i\left(\cdot\right)$ refers to cropping the $i$-th image patch from image $z_t$.

            Subsequently, each window undergoes independent denoising according to \eqref{eq:sd}. The objective of MultiDiffuser is to ensure that $\Psi\left(z_t\right)$ aligns with $\Phi\left(x_t^i\right),\ i\in\left[n\right]$. Therefore, the optimization problem is defined as follows:
            
            \begin{equation} 
            \label{eq:md}
            \begin{aligned}
                z_{t-1}  =\underset{\hat{z}_{t-1}}{\operatorname{argmin}} \mathcal{L}_{MD}\left(\hat{z}_{t-1} \mid z_{t} \right), 
            \end{aligned}
            \end{equation}

            \begin{equation} 
            \label{eq:L-md}
            \begin{aligned}
                \mathcal{L}_{MD}
                 = \sum_{i=1}^{N} W_{i} \otimes \left\|F_{i}\left(\hat{z}_{t-1}\right)-              \Phi\left(F_{i}\left(z_{t}\right)\right)\right\|^{2},
            \end{aligned}
            \end{equation}
            \noindent 
            where $\hat{z}_{t-1}$ represents the variable to be determined in relation to $z_{t}$ and $W_i$ denote the weight matrix of the $i$-th window. 
            
            The framework employs a global least squares optimization to integrate the denoising results across all windows. The final image is then generated through weighted averaging:
            
            \begin{equation} 
            \label{eq:W}
                z_{t-1}=\ \frac{\sum_{i}{W_i\otimes F_i^{-1}\left(\Phi\left(x_t^i\right)\right)}}{\sum_{i} W_i},
            \end{equation}
            \noindent 
            where $F_i^{-1}\left(\cdot\right)$ is the inverse function of $F_i\left(\cdot\right)$.

    \begin{figure*}[t]
    \centering
    \includegraphics[width=1\textwidth]{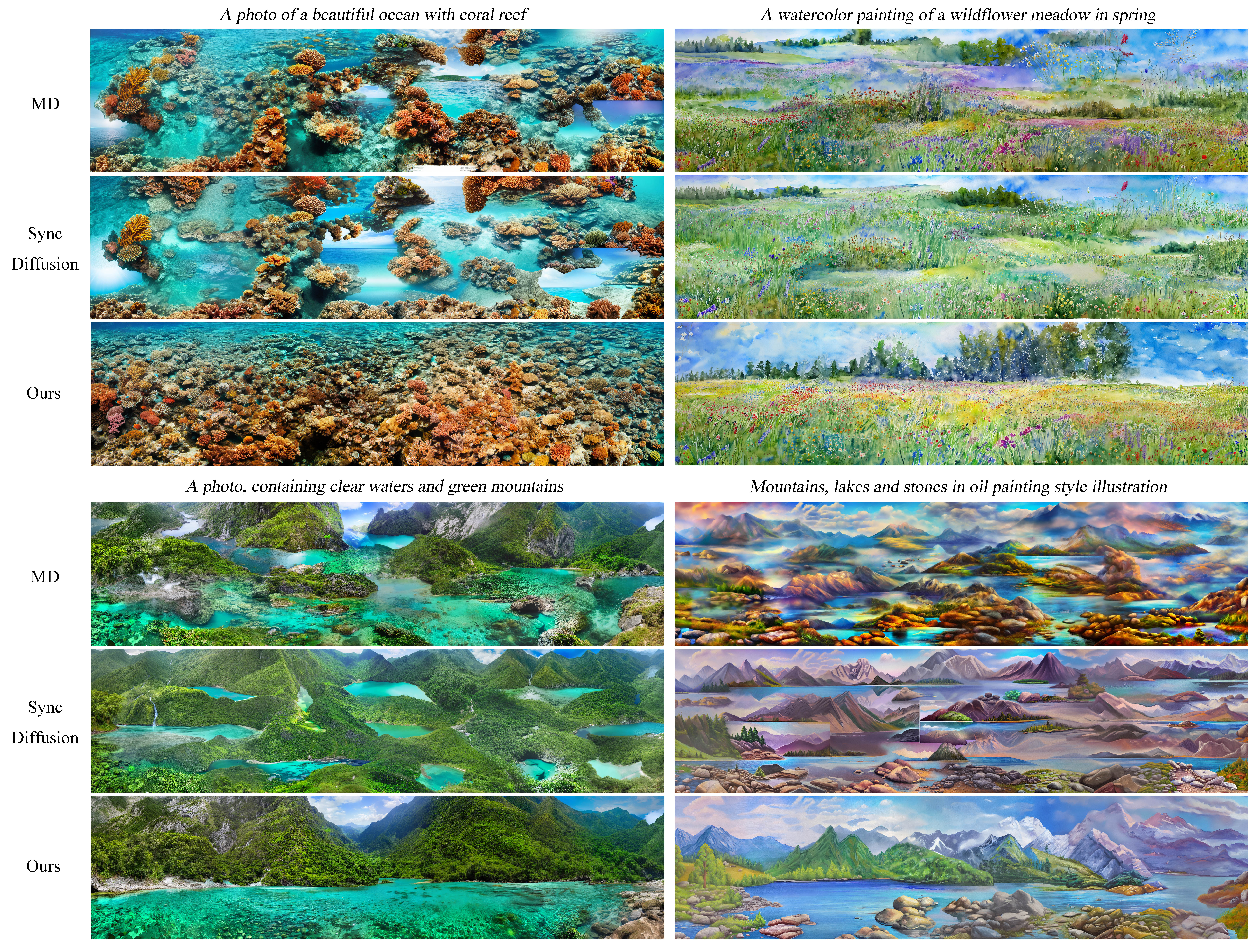} 
    \caption{Qualitative comparisons among MultiDiffusion, SyncDiffusion, Multi-Scale Diffusion. Our approach significantly improves spatial layout issues in high-resolution panoramic generation, producing semantically coherent and visually consistent results.}
    \label{fig:3}
    \end{figure*}

    \subsection{Multi-Scale Diffusion}
    \label{Multi-Scale Diffusion}
        MultiDiffusion effectively generates coherent panoramic images when limited to unidirectional expansion. However, the model exhibits significant limitations when attempting bidirectional expansion, manifesting in disorganized spatial arrangement and visual inconsistencies. These issues arise from the independent generation of panorama segments, which leads to misaligned features and repetitive patterns, as shown in Fig. \ref{fig:1}. A lack of contextual awareness between different segments compromises spatial coherence, diminishing the panorama's realism. This limitation highlights the need for an integrated approach to multi-dimensional panoramic generation.

        To address this limitation, we propose a Multi-Scale Diffusion model that generates coherent, highly detailed panoramic images, as illustrated in Fig. \ref{fig:2}. This framework integrates with existing joint diffusion systems. By extending the MultiDiffusion approach across multiple resolution layers, our model balances structural elements from lower resolutions with fine details from higher resolutions, enhancing overall image quality. We formulate the optimization problem as:

        \begin{equation} 
        \label{eq:5}
        \begin{aligned}
            z_{t-1}^{s}         &=\underset{\hat{z}_{t-1}^{s}}{\operatorname{argmin}} \mathcal{L}_{MSD}\left(\hat{z}_{t-1}^{s} \mid z_{t}^{s}, z_{t-1}^{s-1}\right) \\
                                &=\underset{\hat{z}_{t-1}^{s}}{\operatorname{argmin}}  \mathcal{L}_{MD}\left(\hat{z}_{t-1}^{s} \mid z_{t}^{s}\right) +
                                \omega\mathcal{L}_{MS}\left(\hat{z}_{t-1}^{s} \mid z_{t-1}^{s-1}\right),
        \end{aligned}
        \end{equation}
    
        \noindent 
        where $z_t^s$ denotes the noise image at the $s$-th resolution, $\omega$ is the weight. The equation comprises two components: $\mathcal{L}_{MD}$ represents the optimization objective of MultiDiffusion \eqref{eq:L-md}, while $\mathcal{L}_{MS}$ ensures minimal deviation between denoised images across different resolution layers. The formula is as follows:
    
        \begin{equation} 
        \label{eq:L-ms}
        \begin{aligned}
            \mathcal{L}_{MS} = \sum_{i=1}^{N} W_{i} \otimes \left\|d s\left(F_{i}\left(\hat{z}_{t-1}^{s}\right)\right)-F_{i}^{\prime}\left(z_{t-1}^{s-1}\right)\right\|^{2},
        \end{aligned}
        \end{equation}
        \noindent 
        where $ds\left(\cdot\right)$ refer to the downsampling function (e.g., Bilinear Interpolation), and $F_i^\prime\left(\cdot\right)$ is the crop function associated with $F_i\left(\cdot\right)$ that samples the corresponding region on the low-resolution panoramic image $z_{t-1}^{s-1}$. In contrast to MultiDiffusion, which derives an analytical solution for $z_{t-1}^s$ by averaging features in overlapping regions, we employ backpropagation of gradients to obtain an approximate solution for $z_{t-1}^s$.
        
        In the single-step denoising process of the noisy image $z_t^S$, we create a sequence of lower-resolution images through progressive downsampling, with $z_t^0$ representing the lowest resolution. The denoising process is divided into $S$ stages, where Multi-Scale Diffusion is applied sequentially to the noisy image $z_t^s$ across increasing resolutions, while the initial stage employs MultiDiffusion for denoising.

        At each resolution level $s$ (where $s\textgreater1$), the MSD framework processes images through two parallel paths. In the first path, it applies the cropping function $F_i\left(\cdot\right)$ to the noisy image $z_t^s$ to obtain a window image $x_{t,i}^s$, which is then denoised to produce $\Phi(x_{t,i}^s)$. In the second path, it applies a different cropping function $F_i^\prime\left(\cdot\right)$ to the low-resolution panoramic image $z_{t-1}^{s-1}$, generating the corresponding window image $x_{t-1,i}^{s-1}$.

        Theoretically, the window image $\Phi(x_{t,i}^s)$ that is first denoised and then downsampled should closely resemble the window image $x_{t-1,i}^{s-1}$ that is first downsampled and then denoised. The framework employs the MSE loss between these two window images as an objective function to measure their correspondence. The gradient calculated from this objective function is then used in backpropagation to update the original window image $x_{t,i}^s$. The mathematical formulation is provided below:
    
        \begin{equation} 
        \label{eq:7}
            \hat{x}_{t, i}^{s}=x_{t, i}^{s}-\omega \nabla_{x_{t, i}^{s}}\left\|d s\left(\Phi\left(x_{t, i}^{s}\right)\right)-x_{t-1, i}^{s-1}\right\|^{2},
        \end{equation}
    
        \noindent 
        where $\omega$ is the weight of the gradient descent. MultiDiffusion is applied to denoise and merge these images at the end.
        

\section{Experiment}
    \subsection{Experimental Setting}
        \subsubsection{Baselines} We compare our MSD model with the following two baseline models: (1) MultiDiffusion \cite{BarTal2023MultiDiffusionFD}, a special case of our method when the gradient weight $\omega=0$; (2) SyncDiffusion \cite{lee2023syncdiffusion}, an extension of MultiDiffusion that improves global consistency through perceptual loss. For a fair comparison, we maintained consistent hyperparameters across all models and conducted all experiments using an A100 GPU.

        \subsubsection{Experimental Setup} For the reference model, we employ two variants: the Stable Diffusion v2.0 and the Stable Diffusion v1.5. Operating within a latent space of $\mathbb{R}^{4\times64\times64}$, this model generates images of $\mathbb{R}^{3\times512\times512}$. We aligned the crop window size with the model's default resolution and generated panoramic images of resolution $1024\times4096$ ($128\times512$ in the latent space). The panoramic images' height is doubled, and the width is octupled compared to the original dimensions. Sec. \ref{Comparsion} presents results using specific parameters: gradient weight of $\omega=10$ and scaled cosine decay factor of $\left(1+\cos{\left(\frac{T-t}{T}\times\pi\right)}\right)/2$. Sec. \ref{Ablation} provides a detailed analysis of how these parameters affect image generation.

        \subsubsection{Datasets} We constructed our evaluation dataset following standard practices in the field \cite{BarTal2023MultiDiffusionFD, lee2023syncdiffusion, zhou2024twindiffusion}. The evaluation comprised fifteen diverse textual prompts spanning multiple themes and artistic styles. For each prompt, we generated 500 panoramic images and extracted multiple $1024\times1024$ pixel crops to create test datasets compatible with our evaluation metrics. We also developed a reference dataset using a benchmark model, generating 2,000 images per prompt.

    \subsection{Comparsion}
    \label{Comparsion}
        We conducted a comprehensive evaluation of our MSD model against baseline models, including both qualitative and quantitative analyses.
        
        \subsubsection{Qualitative Comparison}
            Fig. \ref{fig:3} compares images generated by our proposed method and two baseline models. While the MultiDiffusion model stitches scenes well, it produces disorganized layouts and unnatural features in high-resolution panoramic images. The SyncDiffusion model improves stylistic consistency but suffers from spatial structure issues like floating mountains and unsupported lakes. These results demonstrate that merely optimizing window connections and maintaining style consistency alone is insufficient for high-quality panoramic image generation. 

            The proposed MSD method demonstrates superior capability in generating high-resolution panoramic images that maintain visual and semantic coherence across diverse textual prompts. The qualitative assessment of generated images reveals a robust correlation between the spatial relationships specified in the input prompts and the resultant visual compositions, demonstrating the model's efficacy in transforming textual guidance effectively into coherent spatial structures.

            \begin{figure}[ht]
            \centering
            \includegraphics[width=1\columnwidth]{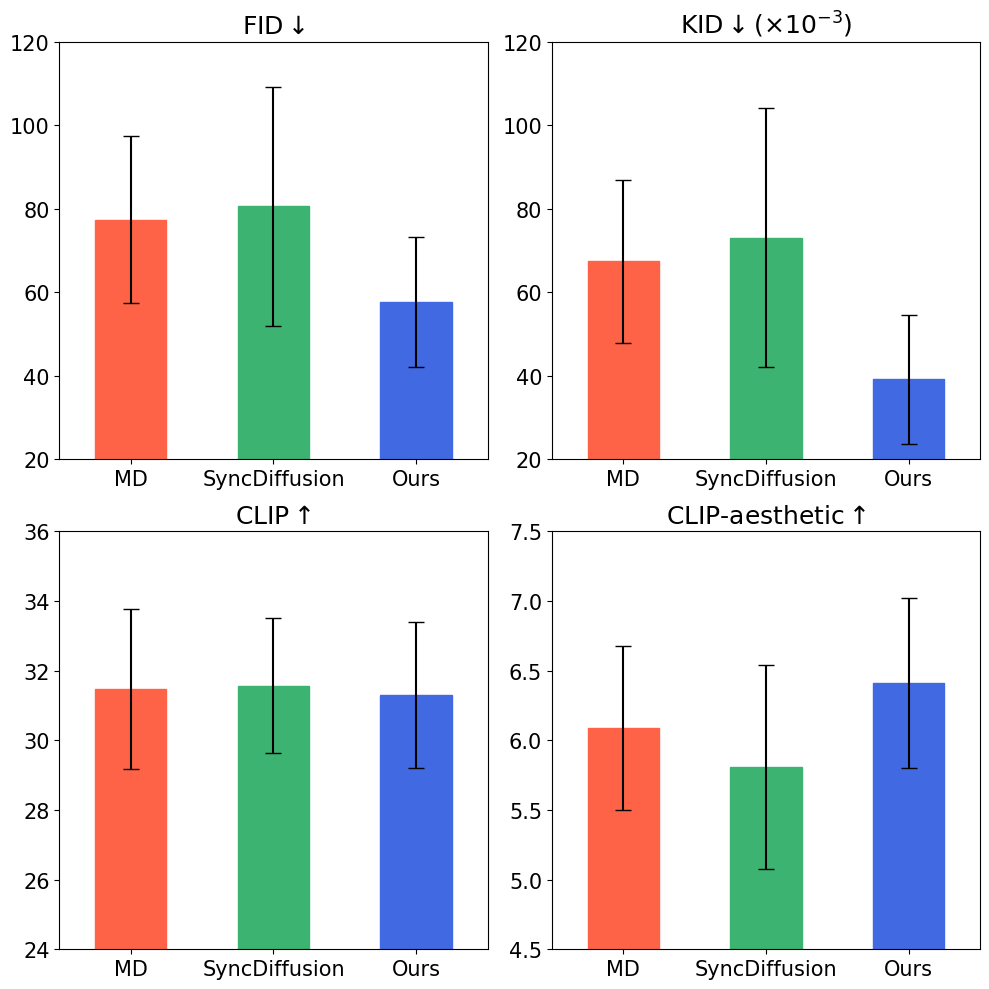} 
            \caption{Quantitative results comparing four key metrics across two dimensions. Our method achieves superior performance, particularly in image quality, establishing it the most effective approach for high-resolution panoramic image generation.}
            \label{fig:4}
            \end{figure}
            
        \subsubsection{Quantitative Comparison}
            We evaluate the generated panoramas using multiple quantitative metrics. These metrics assess the panoramas' fidelity and diversity, as well as their adherence to the input prompts.
            
            \begin{itemize}
                \item \textbf{Fidelity} \& \textbf{Diversity}: \textbf{FID} and \textbf{KID} are employed to assess the fidelity and diversity of the generated images. Both metrics evaluate the distribution of generated images against reference images, with FID calculating feature vector distances and KID utilizing a kernel-based approach. 
                
                \item \textbf{Compatibility}: \textbf{CLIP} is used to evaluate the alignment between generated images and the input prompts by calculating cosine similarity, while \textbf{CLIP-aesthetic} is used to quantify the aesthetic quality of the images using a linear estimator. 
            \end{itemize}

            The quantitative comparison illustrated in Fig. \ref{fig:4} indicates that our MSD approach outperforms the baselines across four critical evaluation metrics. Our method substantially improves panorama image quality, as reflected by lower FID and KID scores, indicating that our generated images more closely align with the reference distribution. The CLIP-Aesthetic scores further corroborate these improvements, with our method achieving the highest ratings. Moreover, our method maintains strong prompt adherence, performing on par with existing methods in terms of CLIP metrics. These comprehensive results demonstrate that our method effectively balances aesthetic quality with prompt relevance, suggesting its viability for high-fidelity panorama generation tasks.

    \subsection{Ablation Study}
    \label{Ablation}
        The gradient weight $\omega$ in \eqref{eq:7} is crucial for optimizing the spatial layout of high-resolution panoramic images. This parameter controls how spatial information from low-resolution images influences their high-resolution counterparts, directly affecting the final image quality. Our experiments revealed significant improvements in FID scores. Given the importance of FID in assessing image quality, we focused our evaluation primarily on this key metric.

\begin{figure}[t]
            \centering
            \includegraphics[width=1\columnwidth]{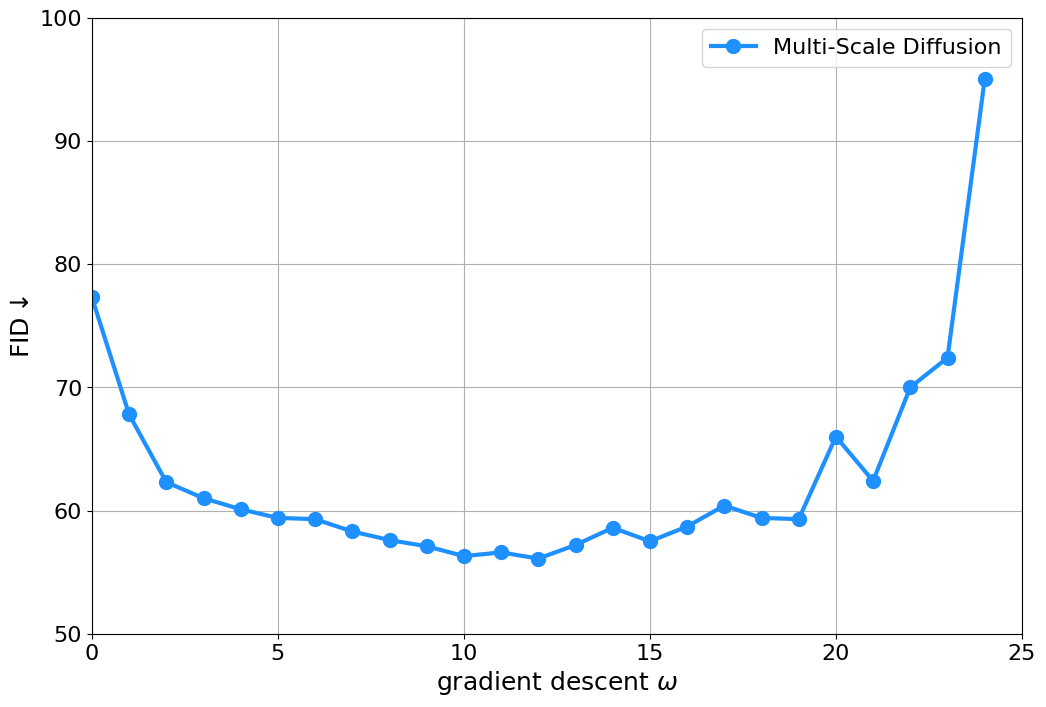} 
            \caption{Impact of gradient weight $\omega$ on fidelity and diversity in high-resolution image generation. Optimal balance is achieved when $\omega=10$. Lower $\omega$ values result in insufficient spatial layout optimization, while higher values may lead to image structure collapse.}
        \label{fig:5}
        \end{figure}

        Fig. \ref{fig:5} illustrates the impact of varying the gradient weight $\omega$ from 0 to 24. At values of $\omega\le2$, the FID metric shows a rapid decrease, indicating significant improvements in image layout. When $\omega \geq 20$ increases the FID metric, suggesting that excessive optimization leads to degraded image generation and reduced quality. We found that $\omega = 10$ yields optimal results, producing high-resolution images that balance spatial layout and fine detail best.
            
\section{Conclusion}
    The Multi-Scale Diffusion is a versatile framework that enhances the generation of high-resolution panoramic images. By operating across multiple resolution levels, it uses information from lower-resolution images to guide the refinement of higher-resolution outputs through gradient descent optimization. This approach produces panoramas that maintain both global structural coherence and fine-scale detail. Our empirical evaluations demonstrate that MSD performs better than baseline methods in quantitative and qualitative assessments. Beyond static images, the model shows potential for video generation, which we aim to explore in future research.

\bibliographystyle{IEEEbib}
\bibliography{MSD}

\onecolumn
\section{Appendix}     
    \subsection{Algorithm}
        As described Section III-B , we introduce the Multi-Scale Diffusion. Here we try to provide the algorithms in details.

        \begin{algorithm}
            \label{alg1}
        
            \caption{Pseudocode of one-time denoising in Multi-Scale Diffusion.}
            \textbf{Input}: \qquad\quad $z_t^S$ \Comment {Noisy images at timestep $t$}\\
            \textbf{Parameter}: \quad$\omega$    \Comment {Gradient descent weight}\\
            \textbf{Output}: \qquad$\ z_{t-1}^S$    \Comment {Noisy images at timestep $t-1$}
            
            \begin{algorithmic}[1]
                \Function {Multi-Scale Diffuser}{$z_t^s,\ z_{t-1}^{s-1}$}
                    \For{$i = 1 \to N$}
                        \State $x_{t,i}^s \gets F_i(z_t^s)$     \Comment{Crop window from panoramic image}
                        \State $x_{t-1,i}^{s-1} \gets F_i^\prime(z_{t-1}^{s-1})$   
                        \State $\hat{x}_{t, i}^{s}=x_{t, i}^{s}-\omega \nabla_{x_{t, i}^{s}}\left\|d s\left(\Phi\left(x_{t, i}^{s}\right)\right)-x_{t-1, i}^{s-1}\right\|^{2}$      \Comment{Gradient descent (Eq. 9)}
                    \EndFor
                    \State \Return{$\left\{{\hat{x}}_{t,i}^s\right\}$}
                \EndFunction
            \end{algorithmic}

            \begin{algorithmic}[1]
                \Function{Denoising One-Step}{$z_t^S$}
                    \For{$s = S \to 2$} \Comment{Downsample}
                        \State $z_t^{s-1}\gets ds\left(z_t^s\right)$
                    \EndFor
                    
                    \State $z_{t-1}^1\gets\frac{\sum_{i}{W_i\otimes F_i^{-1}\left(\Phi\left(F_i\left(z_t^1\right)\right)\right)}}{\sum_{i} W_i}$                \Comment{Apply MultiDiffusion at the lowest-resolution layer (Eq. 6)}

                    \For{$s = 2 \to S$}
                        \State $\left\{{\hat{x}}_{t,i}^s\right\}\gets$ \text{MULTI-SCALE DIFFUSER}$\left(z_t^s,{z}_{t-1}^{s-1}\right)$
                        \State $z_{t-1}^s\gets\frac{\sum_{i}{W_i\otimes F_i^{-1}\left(\Phi\left({\hat{x}}_{t,i}^s\right)\right)}}{\sum_{i} W_i}$        \Comment{Merge images}
                    \EndFor
                    
                    \State \Return{$z_{t-1}^S$}
                \EndFunction
            \end{algorithmic}
        \end{algorithm}

\newpage

     \subsection{More Qualitative Results}
        In this section, we provide more qualitative comparisons between MultiDiffusion, SyncDiffusion, and our Multi-Scale Diffusion on Stable Diffusion v2.0 and v1.5.

            \begin{figure*}[h]
            \label{appendix_1}
                \centering
                \includegraphics[width=0.95\textwidth]{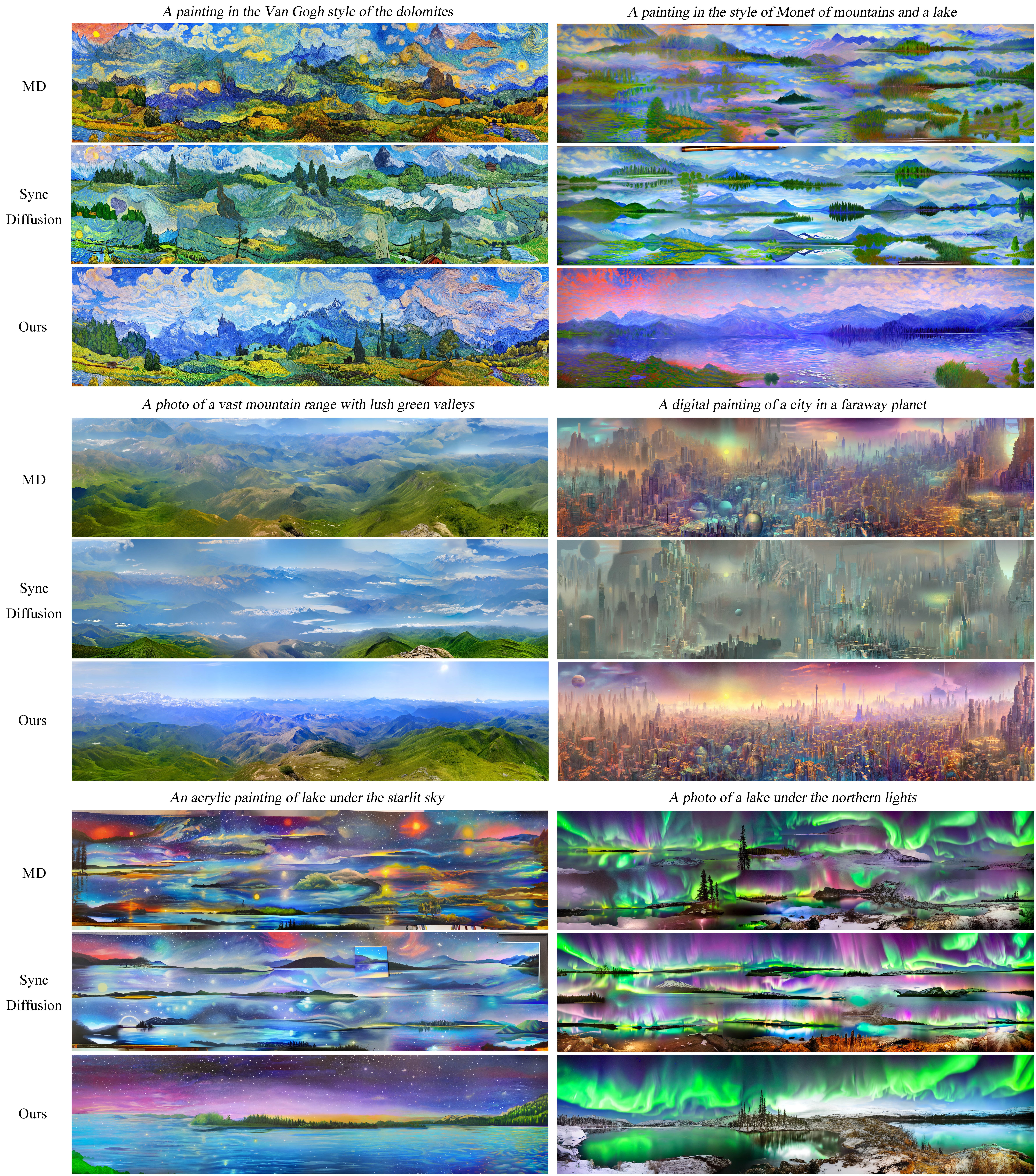} 
                \caption{Additional qualitative comparison results for panorama generation using various text prompts on Stable Diffusion v2.0.}
            \end{figure*}

            \begin{figure*}[h]
            \label{appendix_2}
                \centering
                \includegraphics[width=0.95\textwidth]{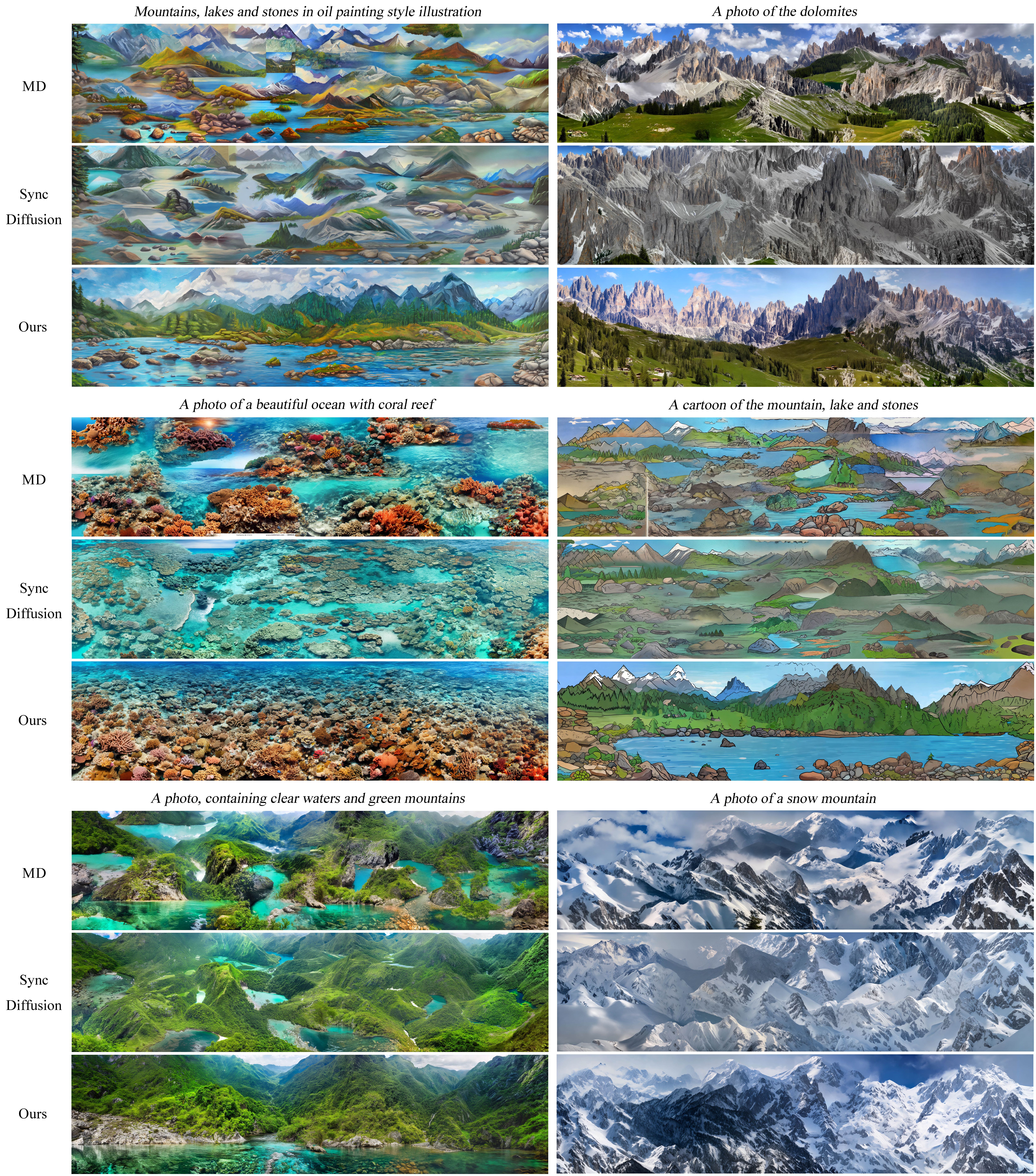} 
                \caption{Additional qualitative comparison results for panorama generation using various text prompts on Stable Diffusion v1.5.}
            \end{figure*}

\end{document}